\documentclass{article}
\usepackage{spconf,amsmath,epsfig}
\usepackage[misc]{ifsym}
\usepackage{multirow}
\usepackage{amssymb}
\usepackage{algorithm}
\usepackage{algorithmic}
\usepackage{float}
\usepackage[table,xcdraw]{xcolor}

\title{SPATIAL-SPECTRAL HYPERSPECTRAL IMAGE CLASSIFICATION VIA MULTIPLE RANDOM ANCHOR GRAPHS ENSEMBLE LEARNING}
\name{Yanling Miao, Qi Wang\sthanks{Corresponding Author.}, Mulin Chen, Xuelong Li}
\address{School of Computer Science and School of Artificial Intelligence, Optics and Electronics (iOPEN), \\ Northwestern Polytechnical University, Xi'an 710072, Shaanxi, P.R. China.}
\begin{document}
%
\maketitle
\begin{abstract}
Graph-based semi-supervised learning methods, which deal well with the situation of limited labeled data, have shown dominant performance in practical applications. However, the high dimensionality of hyperspectral images (HSI) makes it hard to construct the pairwise adjacent graph. Besides, the fine spatial features that help improve the discriminability of the model are often overlooked. To handle the problems, this paper proposes a novel spatial-spectral HSI classification method via multiple random anchor graphs ensemble learning (RAGE). Firstly, the local binary pattern is adopted to extract the more descriptive features on each selected band, which preserves local structures and subtle changes of a region. Secondly, the adaptive neighbors assignment is introduced in the construction of anchor graph, to reduce the computational complexity. Finally, an ensemble model is built by utilizing multiple anchor graphs, such that the diversity of HSI is learned. Extensive experiments show that RAGE is competitive against the state-of-the-art approaches.
\end{abstract}
\section{Introduction}
\label{sec:intro}
Hyperspectral images (HSI) obtained by hyperspectral imaging spectrometer provides abundant spatial structure and spectral information of the observed objects. With very narrow diagnostic spectral bands, HSI can effectively reflect subtle objects between land cover classes~\cite{8961164}. Therefore, HSI has been used in the real-world applications. Among these, the classification is considered as a fundamental task.
 
A plenty of conventional supervised methods, which only learn from labels, have been developed for HSI classification~\cite{8447427,makantasis2015deep}. Since the data labeling is quite costly, the models (i.e., semi-supervised learning) that utilize labeled and unlabeled data are rather meaningful. Due to the simple and elegant formulation, the semi-supervised learning (SSL) has obtained more attention~\cite{nie2019semi}. Especially, graph-based SSL models belong to convex optimization problem and provide the closed-form solution. However, graph-based SSL usually suffer from the high time complexity with computing the inverse of graph Laplacian~\cite{anis2018sampling}. To address the above problem, the anchor graph (AG)-based approaches have been developed, which exploit a small subset of the whole points defined anchors to construct the large-scale adjacent graph~\cite{he2020fast}. It reduces the computational complexity and effectively processes HSI data. Nevertheless, the previous AG-based works rarely use the spatial features within pixels, which limits their discriminability.

Many attempts have been made in spatial-spectral combined approaches~\cite{7425209,huang2019dimensionality,8880686}. The intraclass spatial-spectral hypergraph was constructed by considering the coordinate relationship and similarity between adjacent samples~\cite{luo2018feature}. These methods usually extract the spatial features based on the coordinate distance metric between neighboring pixels, which fail to discovery the fine feature (i.e., local texture feature). In addition, the aforementioned models are based on a single learner, which are weak in diversity learning. To alleviate this situation, the ensemble learning strategy obtains more attention~\cite{8765736}. However, the AG-based ensemble learning method is rarely considered in HSI classification task.

To overcome the aforesaid shortcomings, this paper proposes a novel multiple Random Anchor Graphs Ensemble learning method (RAGE) for HSI classification. The main contributions can be summarized as follows. 

$\bullet$ To capture the fine feature, we adopt the Local Binary Pattern (LBP) method to extract the texture information as the spatial feature, which can refine the local structures of HSI.

$\bullet$ To effectively handle the high-dimensional and large-scale problem, we propose an anchor graph-based SSL model, which learns the adjacent graph by assigning neighbors adaptively. 

$\bullet$ To obtain the optimal predictive model, we introduce the ensemble learning strategy in the proposed scheme, where multiple anchor graphs are constructed in parallel to ensure the runtime efficiency.     
  
\begin{figure*}[htbp]
\textbf{\centering
	\includegraphics[height=5cm,width=16cm]{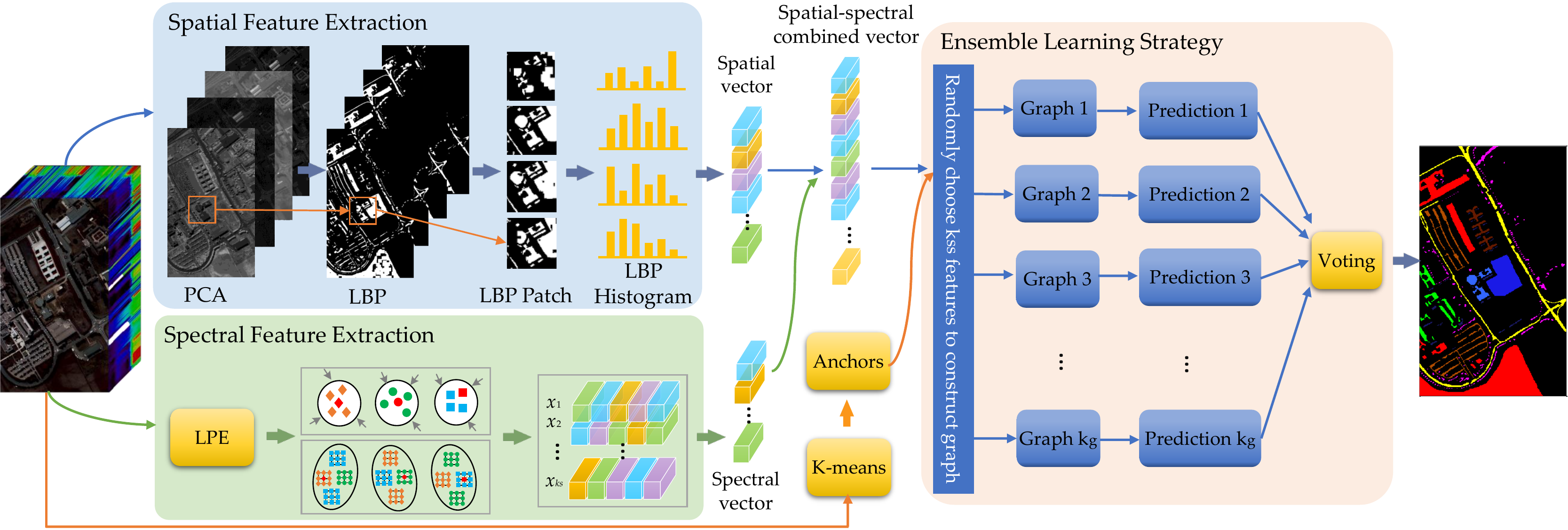}
	\caption{The overview architecture of the proposed RAGE.}
	\label{fig:overview}}
\end{figure*}

\section{METHODOLOGY}

\subsection{Graph-based Semi-supervised Learning Framework}
\label{ssec:gssl}
Graph-based SSL methods utilize the undirected graph to learn the intrinsic geometric structures of HSI data and encode pairwise relations between pixels. 

Suppose that each pixel of HSI can be denoted as $x_{i}\in\mathbb{R}^{d}$ $(i=1,2,...,n)$, where $d$ is the number of spectral bands and $n$ refers to the number of HSI pixels. The HSI data is denoted as $\mathbf{X}=[x_{1},x_{2},...,x_{n}]^{T}\in \mathbb{R}^{n\times d}$. Let $\mathbf{X}=\mathbf{X}_l \cup \mathbf{X}_v (n=l+v)$, where $\mathbf{X}_l=\{x_{1},x_{2},...,x_{l}\}$ is the labeled set with the labels $\mathbf Y(x_{i})\in\{1,2,...,c\}$ and $\mathbf{X}_v=\{x_{l},x_{l+1},...,x_{l+v}\}$ is the unlabeled set. The $ s_{ij} $ represents the similarity between $x_i$ and $x_j$, $\mathbf{S} \in \mathbb{R}^{n \times n}$. 
According to the manifold assumption, the adjacent points $x_i$ and $x_j$ are likely to be divided into the same category if the weight $s_{ij}$ is large. 

For the multi-class classification task, the graph-based SSL methods can be considered as the quadratic optimization problem:
\begin{equation}\label{Eq_1}
\mathop {\min }\limits_{\bf{F}} 
{\mathop {\rm Tr}\limits }((\bf{F-Y})^T\bf{C}(\bf{{F-Y}}))
+{\mathop {\rm Tr}\limits }({{\bf{F}}^T}{\bf{LF}}),
\end{equation}
where $\mathbf{F} \in \mathbb{R}^{n \times c}$ is the predicted labels matrix. $\mathbf{Y}=[y_1,...,y_l,0,...,0]^T \in \mathbb{R}^{n \times c}$ denotes the labels matrix. $\mathbf{C} \in \mathbb{R}^{n \times n}$ is a diagonal matrix whose \textit{i}-th element is represented as $c_i=\alpha_l$ $(1 \leq i \leq l)$ and $c_i=\alpha_v$ $(l+1 < i \leq n)$, where $\alpha_l$ and $\alpha_v$ are two parameters. $\mathbf{L}$ is the Laplacian matrix, where $\mathbf{L=D-S}$ and $\mathbf{D}$ is the degree matrix.

\subsection{Spatial and Spectral Feature Extraction}
\label{ssec:ssf}
The adjacent pixels in a local homogeneous area have the spatial distribution consistency of land objects. 
To refine the spatial information, the LBP model is applied in each band for extracting texture features. The overview architecture of the proposed RAGE is shown in Fig. \ref{fig:overview}. 

Suppose that the neighborhood space $\Omega(x_{i})$ are expressed as $ \{x_{i1},x_{i2},...,x_{i(w^{2}-1)}\}$, where $ w^{2}-1 $ is the number of neighbors of center $ x_{i} $. The local spatial feature of $ x_{i} $ by LBP extractor can be defined as:
\begin{equation}\label{Eq_2}
{\rm LBP}_{w^{2}-1,r}(x_{i})= \sum_{h=1}^{w^{2}-1} L(x_{ih}-x_{i})2^h, x_{ih}\in \Omega(x_{i}),
\end{equation}   
where $r$ is the radius of a circle centered at $x_{i}$. $L(x_{ih}-x_{i}) = 1$ if $x_{ih}>x_{i}$, otherwise $L(x_{ih}-x_{i}) = 0$. As shown in Eq.(2), each neighbor is assigned as a binary label after LBP. It is a non-parametric and simple to calculate. The output is a binary code, which reflects the texture orientation and smoothness within a certain window $w$. Based on the LBP code, the corresponding histogram is obtained over the local patch centered at $x_{i}$. Then all spectral bands of LBP histograms are concatenated to build the spatial feature. 
LBP is insensitive to the monotonic illumination changes, which is suitable for single band processing in HSI.

In spectral feature extraction, we use linear prediction error (LPE) to select spectral bands with distinctive features, which is similar to principal component analysis based on band similarity. Finally, we obtain the spatial-spectral features by stacking the above LBP feature and spectral feature into one-dimensional vector. 

\subsection{Anchor-based Adjacent Graph Construction}
\label{ssec:AG}
After obtaining the spatial-spectral features, we randomly choose $k_{ss}$ features to build an anchor graph in this section.
To reduce the computational complexity, anchor-based strategy is adopted to learn the adjacent graph. 
Ideally, each sample can be represented by the linear combination of the anchors.
Furthermore, the data labels are also the specific representations of the samples.
Therefore, the label prediction function is denoted as
\begin{equation}\label{Eq_3}
f(x_{i})= \sum_{i=1}^{m} P_{ij}f(u_{j}),
\end{equation}
where $ m $ is the number of anchors ($m\ll n$). $ \mathbf{U}=[u_1,u_2,...u_m]^T\in \mathbb{R}^{m\times d} $ denotes the anchor set. $ \mathbf{P} $ $\in \mathbb{R}^{n\times m}$ is the adjacent graph, where $ p_{ij} $ denotes the similarity between $ x_i $ and $ u_j $. Assume that $ \mathbf{F} =[f(x_1),f(x_2),...,f(x_n)]^T $ $ \in \mathbb{R}^{n \times c} $ and $ \mathbf{F}_u =[f(u_1),f(u_2),...,f(u_m)]^T \in \mathbb{R}^{m \times c} $, the Eq. (3) can be rewritten as $\mathbf{F}=\mathbf{P}\mathbf{F}_u$. Therefore, the design of $\mathbf{P}$ is a key problem, which mainly consists of two steps:

\textbf{Anchors Generation:} The anchors are usually generated by $k$-means or random sampling. To yield more representative anchors, we employ $k$-means method.

\textbf{Adjacent Graph Learning:} The matrix $\mathbf{P}$ is constructed by $k$-nearest neighbors method. Inspired by~\cite{nie2019semi,2017A}, we adopt an adaptive neighbors assignment strategy. The nearest anchors assignment of $ x_i $ can be seen as solving the objective function: 
\begin{equation}\label{Eq_4}
\min_{p_i^T \textbf{1}=1, 0 \leq p_{ij} \leq 1} 
\sum_{j=1}^{m} {\|x_i-u_j\|_2^2 p_{ij}} + {\gamma \|p_{ij}\|_2^2},
\end{equation}
where $ p_i^T $ is the \textit{i}-th row of $ \mathbf{P} $. In Eq.(4), the first part is the smoothness term, which ensures the nearby pixels belong to similar semantic labels. The second is the regularization term, which is to prevent the trivial solution of Eq.(4), where $ \gamma>0 $ is a regularization parameter. Let us define $e_{ij}=\|x_i-u_j\|_2^2$, then the objective function (4) can be converted into
\begin{equation}\label{Eq_5}
\min_{p_i^T \textbf{1}=1, 0 \leq p_{ij} \leq 1}  \frac{1}{2} \Big \|p_i+\frac{1}{2 \gamma}e_i \Big \|_2^2.
\end{equation}

According to the convex optimization of the Lagrangian function, we have $ \gamma=\frac{k}{2}e_{i,k+1}-\frac{1}{2} \sum_{j=1}^{k}e_{ij} $, where $ k $ is the number of nonzero values. By this way, we achieve the optimal solution $ p_{ij}^* $ as follows:
\begin{equation}\label{Eq_6}
p_{ij}^*=\frac{e_{i,k+1}-e_{ij}}
{ke_{i,k+1}- \begin{matrix} \sum_{j'=1}^{k} e_{ij'} \end{matrix}}.
\end{equation}

After getting the adjacent graph $\mathbf{P}$, the normalized adjacent graph $ \mathbf{S} $ can be computed as $ \textbf{S} = \textbf{P} \boldsymbol{\Lambda}^{-1} \textbf{P}^T $, where $ \boldsymbol{\Lambda} $ is a diagonal matrix whose \textit{j}-th element is represented as $ \Lambda_{jj}=\begin{matrix} \sum_{i=1}^{n} p_{ij} \end{matrix} $, and $ \boldsymbol{\Lambda} \in \mathbb{R}^{m \times m} $. Intuitively, the element $ s_{ij} $ of matrix $ \mathbf{S} $ is expressed as $ s_{ij}= p_i^T \Lambda^{-1} p_j $ that satisfies $ s_{ij}=s_{ji} $. Moreover, it is easy to prove that matrix $ \mathbf{S} $ is positive semidefinite and doubly stochastic. The above property of $\mathbf{S}$ is crucial for optimizing AG-based SSL model, and the details will be given later. 

\subsection{AG-based SSL Model}
\label{ssec:AGSSL}
The graph-based SSL objective function has described in Eq. (1). As mentioned above, the prediction function of labels is represented as $\mathbf{F}=\mathbf{P}\mathbf{F}_u$. To infer the labels of unlabeled samples in HSI classification, the AG-based SSL model based on section \ref{ssec:AG} can be considered as the following problem:
\begin{equation}\label{Eq_7}
\mathcal{L}({\mathbf F}_u)=
{\mathop {\rm Tr} }\big (({\bf F}_l-{\bf Y}_l)^T({\bf F}_l-{\bf Y}_l)\big )
+\alpha{\mathop {\rm Tr}}({{\bf{F}}^T}{\bf{LF}}),
\end{equation} 
where $ {\mathbf F}_l = {\mathbf P}_l\mathbf{F}_u $, the $ {\mathbf F}_l $ and $ {\mathbf P}_l $ are the sub-matrix of $ \mathbf{F} $ and $ \mathbf{P} $, respectively. $ {\bf Y}_l $ is the labels of labeled samples, $ \| \cdot \|_F $ is the Frobenius norm. The $\alpha$ is the regularization parameter. The Eq. (7) can be represented as
\begin{equation}\label{Eq_8}
\begin{aligned}
\mathcal{L}({\mathbf F}_u)
&=\|{\bf P}_l{\bf F}_u - {\bf Y}_l \|_F^2+\alpha{\mathop {\rm Tr}}({({\mathbf{P}\mathbf{F}_u})^T}\bf {L}({\mathbf{P}\mathbf{F}_u}))\\ 
&=\|{\bf P}_l{\bf F}_u - {\bf Y}_l \|_F^2+\alpha {\mathop {\rm Tr}}({\mathbf{F}_u}^T {\bf L}_A {\mathbf{F}_u}),
\end{aligned}
\end{equation}
where $ {\bf L}_A={\bf P}^T {\bf L} {\bf P} $, and ${\bf L}={\bf D}-{\bf S}$. According to the aforementioned property of $\mathbf{S}$, we have
\begin{equation}\label{Eq_9}
d_{ii}= \sum_{j=1}^{n} s_{ij} = \sum_{j=1}^{n} p_i^T \Lambda^{-1} p_j = p_i^T \sum_{j=1}^{n} \Lambda^{-1} p_j = p_i^T \textbf{1} = 1.
\end{equation} 
Therefore, according to $ \mathbf{L=D-S=I-S}=\bf{I} - \bf{P} \boldsymbol{\Lambda}^{-1} \bf{P}^T$, we can obtain
\begin{equation}\label{Eq_10}
{\bf L}_A={\bf P}^T (\bf{I}- \bf{P} \boldsymbol{\Lambda}^{-1} \bf{P}^T) {\bf P}={\bf P}^T{\bf P}-({\bf P}^T{\bf P})\boldsymbol{\Lambda}^{-1}({\bf P}^T{\bf P}).
\end{equation} 
The optimization problem (8) can be solved by setting the first derivative as zero, and the final solution is
\begin{equation}\label{Eq_12}
{\bf F}_u^*=({\bf P}_l^T{\bf P}_l+\alpha{\bf L}_A)^{-1}{\bf P}_l^T{\bf Y}_l.
\end{equation}

After obtaining the $ {\bf F}_u^* $, the predicted label of unlabeled pixel $ x_i $ can be determined by
\begin{equation}\label{Eq_13}
y_i= \arg \mathop {\max }\limits_{j \in \{1,...,c\}} {\bf P}_{i.}{\bf F}_{u_j}^*.
\end{equation}

Actually, we will build $ k_g $ graphs by repeating section \ref{ssec:AG}. Based on ensemble learning strategy, the labels of unlabeled samples are determined by the voting results of $ k_g $ graphs acquired from section \ref{ssec:AGSSL}. 
%
\section{EXPERIMENTS}

\subsection{Datasets and Experimental Setup}
To demonstrate the effectiveness of the proposed RAGE method, the experiments are conducted on public Indian Pines and Pavia University datasets. Indian Pines dataset contains 16 land cover types, which leaves 200 bands to be used for experiments after removing noise and water absorption bands. Pavia University includes 9 classes and 103 bands for classification task. The proposed RAGE is compared with R-VCANet~\cite{makantasis2015deep}, LSLRR~\cite{8447427}, SS-RMG~\cite{gao2018spectral} and SSHGDA~\cite{luo2018feature}. The parameters of RAGE are $\{\alpha=\alpha_l=0.01, \alpha_v=10^{-6}, w=7, k_{ss}=96, k_g=4 \}$. We randomly choose 5\% and 1\% of Indian Pines and Pavia University datasets for training, respectively. Three quantitative metrics are used to evaluate performance, including overall accuracy (OA), average accuracy (AA) and Kappa coefficient.
\begin{figure}[htb]
	\vspace{-0.3cm}
	\textbf{\centering
		\includegraphics[height=4cm,width=8.5cm]{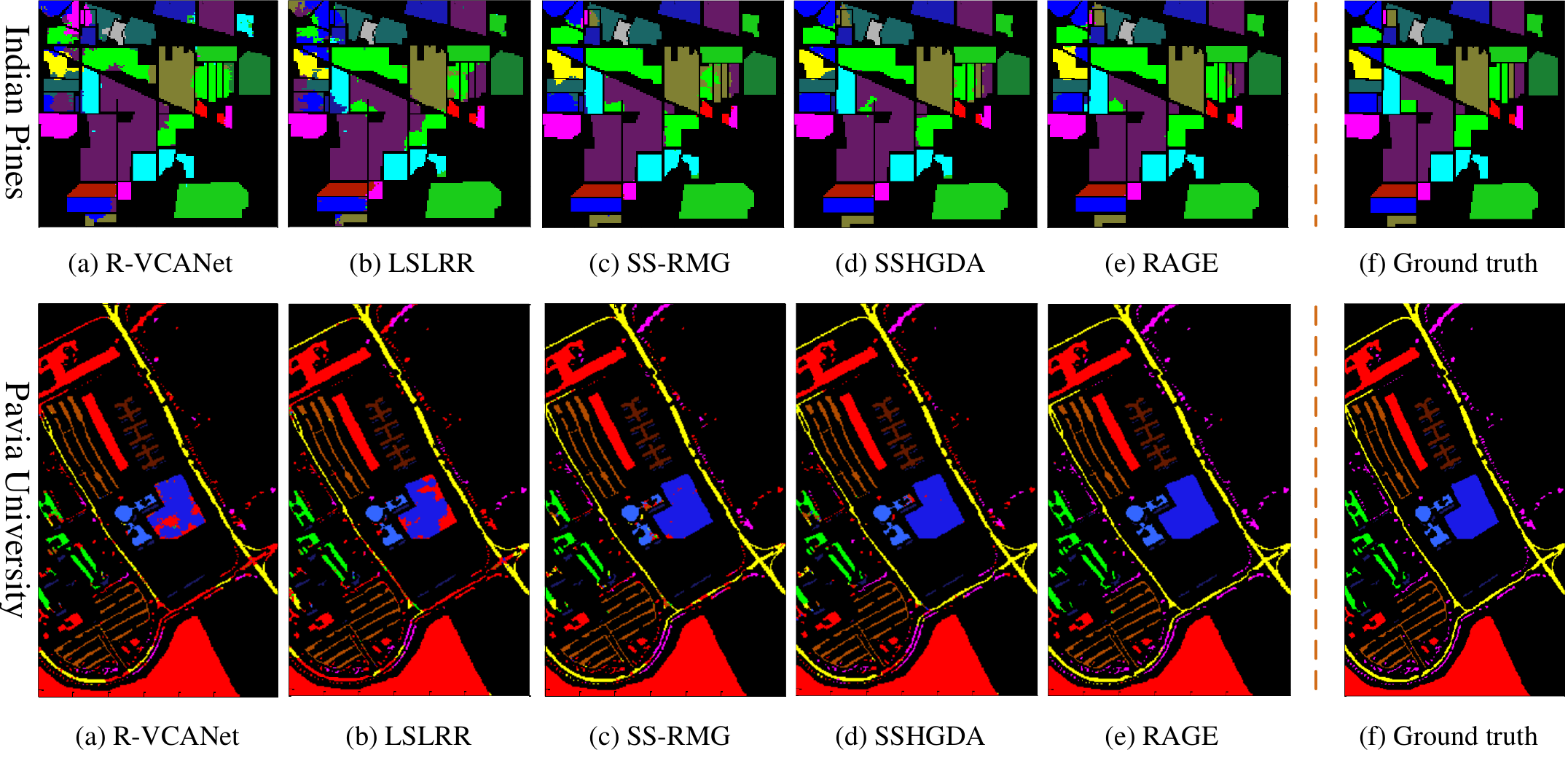}
		\caption{The classification maps of different methods.}
		\setlength{\abovecaptionskip}{0pt}
		\setlength{\belowcaptionskip}{0pt}
		\label{fig:result}}	
	\vspace{-1.7em}
\end{figure}  

\subsection{Experimental Results and Analyses}
\begin{figure}[htb]
	\textbf{\centering
		\includegraphics[height=3cm,width=6cm]{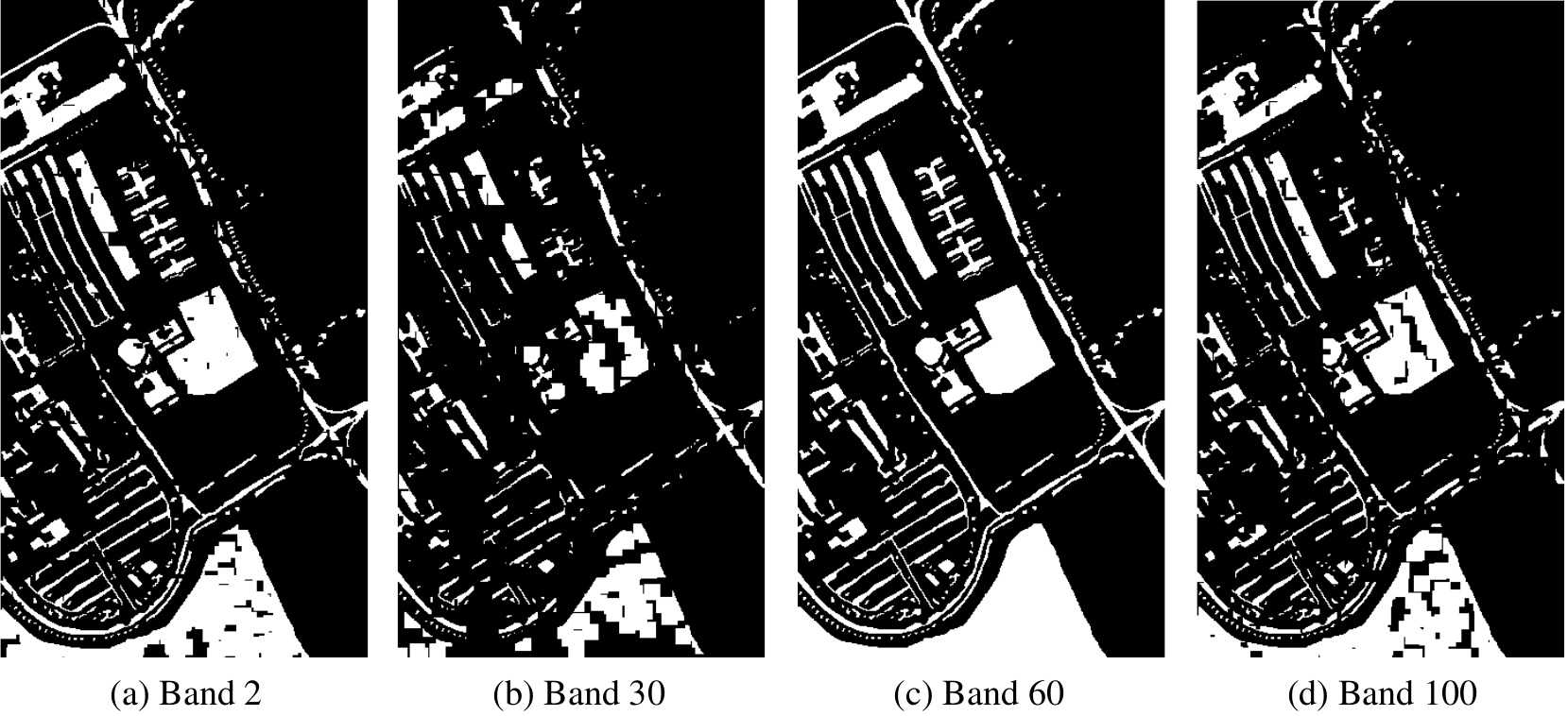}
		\setlength{\abovecaptionskip}{0pt}
		\setlength{\belowcaptionskip}{0pt}
		\caption{The spatial features based on LBP on bands.}
		\label{fig:LBP}}
	\vspace{-1.5em}	
\end{figure}

The visual and quantitative results of the methods are given in Fig. \ref{fig:result}, Table \ref{tab1} and Table \ref{tab2}. It can be revealed that RAGE obtains better classification results and the higher accuracy than competitors from the perspective of OA, AA and Kappa. By adopting LBP feature extractor, the distinctive spatial features are obtained as shown in Fig. \ref{fig:LBP}, which discovery the fine differences from different bands and enrich the local structure information. This enhance the discriminability of learners and effectively improve the classification performance of the proposed RAGE. Moreover, we randomly choose several subsets of features to build different anchor graphs in parallel, then obtaining the best prediction from multiple learners by voting. By this way, the classification accuracy and efficiency of the RAGE are further improved with small training samples. From Table \ref{tab1} and Table \ref{tab2}, RAGE consumes the least running time on two datasets, and its accuracy is still maintained at a good level comparing with other algorithms. 
\begin{table}[htb]
	\vspace{-0.5cm}
	\caption{Quantitative metrics of different methods on Indian Pines dataset. The optimal value is highlighted in bold.}
	\label{tab1}
	\small
	\centering
	\begin{tabular}{c|c|c|c|c}
		\hline
		\textbf{Methods}  & \textbf{OA(\%)} & \textbf{AA(\%)} & \textbf{Kappa}  & \textbf{Time(s)} \\ \hline
		\textbf{R-VCANet} & 88.09           & 90.97           & 0.8639          & 1378.5           \\ \hline
		\textbf{LSLRR}    & 90.97           & 92.65           & 0.8966          & 336.2            \\ \hline
		\textbf{SS-RMG}   & 96.89           & \textbf{97.73}  & 0.9646          & 63.9             \\ \hline
		\textbf{SSHGDA}   & 97.78           & 96.22           & 0.9747          & 1029.7           \\ \hline
		\textbf{RAGE}     & \textbf{98.75}  & 96.90           & \textbf{0.9857} & \textbf{14.2}    \\ \hline
	\end{tabular}
	\vspace{-0.9cm}
\end{table} 
\begin{table}[htb]
	\caption{The results of methods on Pavia University dataset.}
	\label{tab2}
		\small
	\centering
	\begin{tabular}{c|c|c|c|c}
		\hline
		\textbf{Methods}  & \textbf{OA(\%)} & \textbf{AA(\%)} & \textbf{Kappa}  & \textbf{Time(s)} \\ \hline
		\textbf{R-VCANet} & 85.78           & 82.31           & 0.8002          & 2789.7           \\ \hline
		\textbf{LSLRR}    & 91.27           & 86.89           & 0.8809          & 1257.3           \\ \hline
		\textbf{SS-RMG}   & 92.71           & 87.73           & 0.9009          & 380.9            \\ \hline
		\textbf{SSHGDA}   & 93.53           & 89.29           & 0.9123          & 13938.1          \\ \hline
		\textbf{RAGE}     & \textbf{99.26}  & \textbf{98.98}  & \textbf{0.9902} & \textbf{179.5}   \\ \hline
	\end{tabular}
	\vspace{-0.7cm}
\end{table}
\section{CONCLUSION}
In this paper, we propose an efficient spatial-spectral HSI classification method based on multiple random anchor graphs ensemble learning. By adopting LBP model, the fine spatial features are obtained, such that the discriminability is enhanced. The multi-graphs with random spatial-spectral features are built in parallel, which further improves the efficiency of the proposed model and learns the diversity of HSI data. The adaptive neighbors assignment in AG construction decreases the computational complexity. Extensive experiments on two public HSI datasets verify the effectiveness and advantages of the proposed RAGE.


%


\small
\bibliographystyle{IEEEbib}
\bibliography{refs}

\end{document}